\documentclass[11pt,a4paper]{article}

\usepackage{amsmath,amsthm}
\usepackage{fontspec}
\usepackage{unicode-math}
\usepackage[margin=0.95in]{geometry}
\usepackage{booktabs,tabularx,array}
\usepackage{graphicx}
\usepackage{xcolor}
\usepackage[hidelinks]{hyperref}
\usepackage[capitalise,noabbrev]{cleveref}
\usepackage{enumitem}
\usepackage[round,authoryear]{natbib}
\usepackage{microtype}
\usepackage{caption}

\setlist{itemsep=2pt,topsep=4pt}
\newtheorem{proposition}{Proposition}
\theoremstyle{definition}
\newtheorem{definition}{Definition}

\newcommand{\E}{\mathbb{E}}
\newcommand{\qd}{q^{\dagger}}

\title{\textbf{Counting the Bandages}\\[6pt]\large On measuring the time that AI saves}
\author{Solomon Shalom Li{\kern0pt}jo\\ Independent Researcher\\ \texttt{solomon@getromy.app}}
\date{October 2026}

\begin{document}
\maketitle

\begin{abstract}
\noindent I consider what it means for AI assistance to save time. I contrast a stopwatch view, on which time saved is whatever the clock records, with a ledger view, on which it is a counterfactual that has to be read next to the errors the work lets through, and I argue for the latter. The argument has four parts. I define a model-free pair of quantities, time saved and excess escaped defects, and one decision number built from them: the break-even defect cost. I give a baseline model of checking in which, until the model is good enough not to check, a calibrated worker's escaped-defect rate does not depend on how good the model is; checking time grows with the logarithm of the model's error rate; and over-trust buys minutes on the clock at a larger cost in total minutes. I show what randomized studies can and cannot identify, and I replicate METR's developer study from its public data: the same 246 issues support a slowdown of 19\% or 24\% depending on the estimand (34\% without adjusting for a chance imbalance), and AI helped on somewhere between 3\% and 82\% of them. Finally, I restate the published record, with stated bounds in place of imported point assumptions. In GDPval's own review-every-output workflow, review time alone caps GPT-5's speedup at $3.6\times$; the consultants outside the frontier were faster, and more often wrong. Last, I test two of the model's predictions in a small pre-specified experiment on AI reviewers checking AI drafts of the same kind of brief, built from SEC filings made after the models' training data were collected. Told the drafter's true error rate, the reviewer checked more as it rose, but all-or-nothing, switching near where its prompt's costs break even. Told that the drafter erred ten times less often than it did, it checked 7 fewer of 9 figures and let a wrong figure through in 22.5\% of briefs instead of none: faster and worse, entirely through reduced effort. This paper replaces my earlier TSB proposal.
\end{abstract}


% ─────────────────────────────────────────────────────────────────────────────
\section{Two razors}
\label{sec:intro}

Frank Gilbreth, the efficiency engineer, buttoned his waistcoat from the bottom up because it took three seconds instead of seven. For a while he also shaved with two razors at once. He gave that up. ``I can save forty-four seconds,'' he grumbled, ``but I wasted two minutes this morning putting this bandage on my throat'' \citep{gilbreth1948cheaper}.

It is the whole literature on AI-assisted work in miniature, a century early and without the AI. It has a tool that makes the visible part of a job faster, a cost that turns up somewhere else, and a small sample. It also has something the modern literature often lacks. Gilbreth counted the bandages.

The modern literature is large, and it disagrees with itself. Developers using Copilot finished a bounded programming task in 56\% less time \citep{peng2023copilot}. Experienced open-source developers using early-2025 AI tools took 19\% longer \citep{becker2025metr}. Consultants were 25\% faster on tasks inside the frontier of what the model does well; consultants given a task chosen to lie just outside it were faster too, and 19 points less likely to be right \citep{dellacqua2026jagged}. OpenAI's announcement of GDPval said frontier models can complete its tasks ``roughly 100x faster and 100x cheaper than industry experts'' \citep{openai2025gdpvalblog}. The GDPval paper itself, once an expert reviews the output and redoes what fails, puts the figure for GPT-5 at 1.12--1.39$\times$ \citep{patwardhan2025gdpval}.

But what exactly is time saved? There are two ways of thinking about it.

On the first, which I will call the \emph{stopwatch view}, time saved is what a clock records. Time the task with the tool, time it without, subtract. Whatever happens after the clock stops belongs to somebody else's study.

On the second, the \emph{ledger view}, time saved is a counterfactual about two whole workflows, one with the tool and one without, each run until its output is used. The clock records where time was spent. It does not record what the time bought. Some of what looks like saved time is moved time: from writing to reading, from now to later, from me to whoever finds my mistake. On the stopwatch view, checking is a cost of saving time. On the ledger view, not checking is the most common way of saving it.

A thought experiment separates the two. Two analysts get the same model and the same task: a one-page brief with figures from three companies' filings, which takes ninety minutes by hand. The first reads the draft against the filings, finds two wrong figures, fixes them, and is done in thirty-four minutes. The second reads the draft, likes it, and is done in six. Who saved more time? On the stopwatch view the second, by a wide margin; the second analyst has discovered the oldest productivity technology there is, which is not checking. On the ledger view the second analyst has saved nothing yet. They have borrowed twenty-eight minutes against two wrong figures, at a rate that will be set by whoever finds them. That person deserves our respect as well as our sympathy.

I argue for the ledger view, and I try to make it something one can measure. The paper has five parts.
\begin{enumerate}[leftmargin=*]
  \item \textbf{An estimand} (\cref{sec:estimand}). Time saved and excess escaped defects, both measured on final deliverables and both model-free; and one number that turns them into a decision, the \emph{break-even defect cost}: the downstream price of an escaped error at which the tool stops paying.
  \item \textbf{A model of checking} (\cref{sec:model}). In a baseline model built on classical search theory, a worker who checks optimally lets defects through at a rate that does not depend on how good the AI is, until the AI is good enough not to check; checking time grows with the logarithm of the AI's error rate; and a worker who over-trusts the AI by a factor $m$ looks faster on the clock while losing $\kappa(m-1-\ln m)$ minutes on the ledger, where $\kappa$ is the mean time it takes to find a defect that is there. The qualitative result, faster and worse, holds for any detection curve.
  \item \textbf{Identification} (\cref{sec:ident}). What randomized studies identify (averages), what they only bound (how often the tool helps), and what they cannot see at all (per-task effects); the winner's curse in ``frontier'' statistics; and what it costs to certify that a model needs no checking.
  \item \textbf{The record, re-read} (\cref{sec:sims,sec:record}). Simulation checks of the estimators and closed forms; an exact replication and re-analysis of METR's public data; and the published headline numbers restated with bounds instead of imported point assumptions.
  \item \textbf{A test} (\cref{sec:reviewers}). A pre-specified experiment in which an AI reviewer checks AI drafts against SEC filings, with exact ground truth, under error rates the experimenter sets. Checking rises with the stated error rate, though all-or-nothing rather than graded, and over-trust is faster and worse.
\end{enumerate}
\Cref{sec:standard} turns this into a reporting checklist, and \cref{sec:borrowed} lists what I borrowed. A good deal is borrowed.

\paragraph{Relation to TSB.} In April I proposed a Time-Saved Benchmark with five named metrics \citep{lijo2026tsb}. It was wrong in ways worth stating plainly. Its reliability-adjusted metric charged a full redo on top of observed completion times that already contained the redo; on observed data it equals true time saved minus the draft error rate. It treated errors that ship as if they cost redo time, when they cost something else. And two of its headline numbers did not survive contact with their sources. This paper replaces it rather than patching it. The corrections are listed, with executable tests, in the repository.

\paragraph{What this paper is not.} It reports no new human-subjects data. Its one experiment (\cref{sec:reviewers}) is on AI reviewers, one model given explicit costs, and says nothing directly about people. The model in \cref{sec:model} is a model, and its predictions are labelled as predictions. The rest of the empirical content is a replication, a re-analysis and a restatement.

% ─────────────────────────────────────────────────────────────────────────────
\section{What to measure}
\label{sec:estimand}

Fix a population of tasks and workers, a tool, and a way of using it. Call the unaided workflow $B$ and the assisted one $A$. Each ends with a deliverable that someone will use. Let $T_A$ and $T_B$ be the worker time each takes, from receiving the task to handing over the deliverable, and let $D_A$ and $D_B$ indicate whether the handed-over deliverable contains a defect that an independent audit would find. A defect is whatever the task's owner would send back: a wrong figure, a failing test, a fabricated citation. The audit is the key instrument. It is applied blind, in both arms, after the worker is done.

\begin{definition}[Time saved and excess escapes]
\[
\tau \;=\; 1-\frac{\E[T_A]}{\E[T_B]}, \qquad \delta \;=\; \Pr(D_A=1)-\Pr(D_B=1).
\]
\end{definition}

Three remarks. First, $\tau$ is a ratio of means: the share of aggregate unaided time that the assisted workflow does not spend. It is what a manager means by ``we saved 30\% of the hours'', and \cref{sec:ident} explains why the alternatives mislead. Second, $\delta$ can be negative. A tool that catches the worker's own mistakes improves quality, and $\delta<0$ says so. Third, neither number depends on any model of how the work gets done. Both are measured.

The pair falls into one of three regions. If $\tau\ge 0$ and $\delta\le 0$, the tool is no slower and no worse: it \emph{dominates}. If $\tau\le 0$ and $\delta\ge 0$, not both zero, it is no faster and no better: it is \emph{dominated}. Otherwise there is a trade-off, and something has to price a defect against a minute.

I do not want to choose that price. The right price of a wrong figure in a board memo is not the right price of a wrong colour on a slide, and any default I picked would be stipulation. (TSB made this argument against composite scores. It was the best thing in that paper.) Instead I report the price at which the decision flips.

\begin{definition}[Break-even defect cost]
With $\bar T_B=\E[T_B]$ and $\delta\ne 0$, the break-even defect cost is $\lambda^\star=\tau\bar T_B/\delta$, in minutes per defect.
\end{definition}

If an escaped defect costs $c$ minutes to deal with downstream, the net saving per task is $\tau\bar T_B-\delta c$. So when $\delta>0$ the tool pays exactly when $c<\lambda^\star$, and when the tool is slower but cleaner ($\tau<0$, $\delta<0$) it pays exactly when $c>\lambda^\star$. A tool that saves 18 minutes per task and lets through three extra defects per hundred tasks pays if and only if a defect costs less than 600 minutes downstream. Readers who know their own stakes can read off the decision without accepting mine.

None of this is new to health economists. $\lambda^\star$ is the incremental cost-effectiveness ratio, with minutes in place of money and escaped defects in place of health outcomes, and it is read the same way: against a threshold, drawn as a line through the origin of the cost-effectiveness plane \citep{black1990ce}. I borrow the idea and its vocabulary. What is new is only using it on time-saved claims, which have reported time and quality separately but never the price at which they trade off.

Money goes in a separate column. Inference has a price per task, and the expected cost of a correct output can be compared across models and against a human expert without converting anything into wages \citep{erol2026costofpass}. I keep minutes and dollars apart because the exchange rate between them is exactly the kind of context a benchmark should not assume.

% ─────────────────────────────────────────────────────────────────────────────
\section{A model of checking}
\label{sec:model}

The estimand needs no model. The model is here to explain why published numbers look the way they do, and to make predictions someone can test.

\paragraph{Setup.} A task takes $H$ minutes unaided. With assistance, the worker spends $P$ minutes specifying it and $L$ minutes waiting, and receives a draft that is defective with probability $q$. They then check it for $V$ minutes. Checking is a search: if a defect is present, the time to find it is exponential with mean $\kappa$, so the chance of finding it within $V$ minutes is
\begin{equation}
  s(V)=1-e^{-V/\kappa}.
  \label{eq:search}
\end{equation}
This is the exponential detection law of search theory: looking steadily, with no better idea where the thing is, the chance of \emph{not yet} having found it decays exponentially \citep[eqs.~3 and 40]{koopman1946search}. A found defect costs $R$ minutes to repair (fix, re-prompt or redo). An unfound one ships and costs $c>R$ minutes downstream, paid by someone, often someone else. The assumptions are: a draft has at most one defect that matters (A1); detection follows \eqref{eq:search} (A2, relaxed below); repaired drafts are sound (A3); the worker chooses $V$ to minimise expected minutes including $c$, using a belief $\hat q$ about the error rate (A4); and the worker commits to checking for $V$ minutes, reading the whole draft rather than stopping at the first problem (A5, relaxed below). Clock time and the escape rate are then
\begin{equation}
  \E[T_A]=P+L+V+q\,s(V)\,R, \qquad \varepsilon=q\bigl(1-s(V)\bigr).
  \label{eq:time}
\end{equation}

\begin{proposition}[Ceiling]\label{prop:ceiling}
For any detection curve and any $q$, if the worker spends $P+L$ minutes specifying and waiting and $V$ minutes checking (in expectation, if checking can stop early), then $\E[T_A]\ge P+L+V$; so the speedup $H/\E[T_A]$ is at most $H/(P+L+V)$.
\end{proposition}

This is not deep. It is Amdahl's law \citep{amdahl1967validity} wearing a lanyard: the part of the work the tool cannot speed up bounds the speedup of the whole. GDPval's review formula contains it as the limit of a perfect model \citep{patwardhan2025gdpval}, and \citet{liang2026novelty} makes the Amdahl analogy explicit. I state it because everything else rests on it, and because accuracy benchmarks do not measure $P+L+V$. On GDPval's own numbers the ceiling for GPT-5 is $3.6\times$ (\cref{sec:record}).

The ceiling holds $V$ fixed. The interesting question is how much a sensible worker checks.

\begin{proposition}[Optimal checking]\label{prop:optimal}
Under A1--A5 with a calibrated belief ($\hat q=q$), let $\qd=\kappa/(c-R)$. Then:
\begin{enumerate}[label=(\roman*),leftmargin=*]
  \item the worker checks for $V^\star=\kappa\ln^{+}(q/\qd)$ minutes;
  \item the escaped-defect rate is $\varepsilon^\star=\min(q,\qd)$;
  \item while $q/10>\qd$, a tenfold fall in $q$ removes $\kappa\ln 10\approx 2.3\kappa$ minutes of checking.
\end{enumerate}
\end{proposition}

Proofs are in \cref{app:proofs}. Parts (i) and (ii) are not new mathematics. They are the one-cell case of Koopman's optimal distribution of search effort under exponential detection, in which effort goes only where the target is likely enough, and every place searched is left with the same chance of a miss \citep{koopman1957search,stone1975theory}. What is new is reading them as a statement about AI. Three readings.

First, $\qd$ is the error rate below which checking does not pay. With $\kappa=5$ minutes, $R=10$ and $c=600$ (a defect that costs ten hours to unwind), $\qd\approx 0.85\%$. A model that is wrong more often than that should be checked. One that is wrong less often should not.

Second, part (ii) is the surprising one. Above $\qd$, a calibrated worker lets defects through at the rate $\qd$, whatever the model's error rate. A better model makes the work faster, not cleaner. Quality improves only once $q$ falls below $\qd$, where the worker stops checking and escapes equal $q$. In the checking regime, one lever that improves quality is $\kappa$ (raising the stakes $c$ or cutting the repair cost $R$ are others): cheaper checking lowers $\qd$, so fewer defects ship, and it lowers total expected minutes (downstream included) whenever $q>\qd$. It shortens the clock too once $q$ is well above the threshold, past $e^{1+R/(c-R)}\qd\approx 2.8\qd$ here; just above the threshold, cheaper checking makes it worth checking more. This is a formal version of the argument that verifiability, more than accuracy, is what makes assistance useful \citep{fok2024verifiability}.

Third, part (iii): leaderboards are linear in accuracy, and checking is linear in log-error. Where the stakes put $\qd$ below $0.1\%$, moving from 90\% to 99\% accuracy and from 99\% to 99.9\% look very different on a leaderboard and buy the same $2.3\kappa$ minutes of checking.

\begin{figure}[t]
  \centering
  \includegraphics[width=\linewidth]{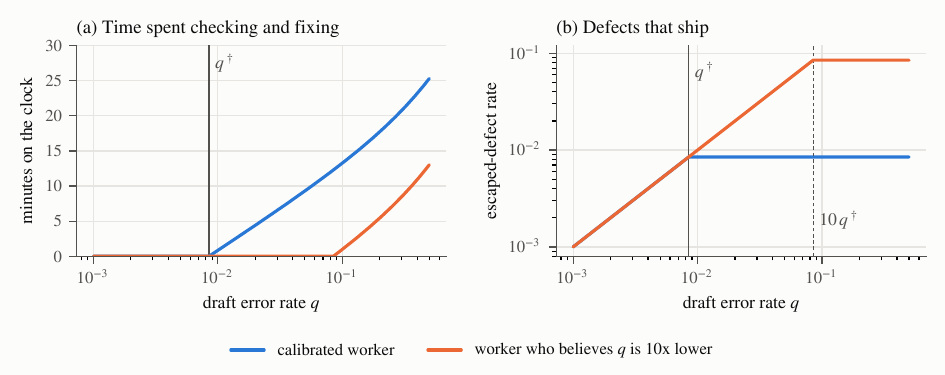}
  \caption{What a worker who checks optimally does as the model improves ($\kappa=5$, $R=10$, $c=600$ minutes, so $\qd\approx 0.85\%$). (a) Checking and repair time is zero below $\qd$ and grows with $\ln q$ above it; the over-trusting worker is never slower. (b) Escaped defects are pinned at $\qd$ for the calibrated worker. The over-trusting worker matches them below $\qd$, lets through everything (escapes equal $q$) between $\qd$ and $10\qd$, and lets through ten times as many above $10\qd$. Generated by \texttt{reproduce.py}.}
  \label{fig:checking}
\end{figure}

\begin{proposition}[Over-trust is faster and worse]\label{prop:overtrust}
Suppose the worker believes the error rate is $\hat q=q/m$.
\begin{enumerate}[label=(\roman*),leftmargin=*]
  \item Under A1 and A3--A5, for \emph{any} detection curve (any distribution of the time it takes to find a defect), the least and the greatest optimal checking times are nondecreasing in $\hat q$. Over-trust ($m>1$) therefore weakly shortens clock time and weakly raises escapes.
  \item Under A1--A5, if $q>\max(1,m)\,\qd$ (so that both workers check something), the belief $q/m$ changes clock time by $-[\kappa\ln m+(m-1)\qd R]$, multiplies escapes by $m$, and raises expected total minutes, downstream included, by exactly
  \[
    \Delta(m)=\kappa\,(m-1-\ln m)\;\ge\;0 .
  \]
  If instead $\qd<q\le m\qd$, the over-trusting worker does not check at all and the loss is $\Delta(q/\qd)\le\Delta(m)$.
\end{enumerate}
\end{proposition}

Part (i) is monotone comparative statics \citep{topkis1978minimizing,milgrom1994monotone}: believed error and checking time are complements, so whatever the shape of $s$, less believed error means less checking. Part (ii) puts numbers on it. With $\kappa=5$ and $m=10$, the over-trusting worker's clock shows about 12 minutes saved per task whenever $q>10\qd$. Their ledger shows 33.5 minutes lost, once the extra defects are paid for. The stopwatch rewards over-trust. The ledger charges for it.

The loss $m-1-\ln m$ is asymmetric. Over-trusting by a factor of ten costs $6.7\kappa$ (when $q>10\qd$); under-trusting by a factor of ten costs $1.4\kappa$ (when $q>\qd$). If you must be wrong about your model, be wrong in the paranoid direction.

This suggests one reading of the jagged frontier. \citet{dellacqua2026jagged} chose, in advance, a task the model handled badly but that looked like ordinary consulting work, and consultants given the model were faster than the control group and more often wrong (\cref{sec:record}). That is the pattern \cref{prop:overtrust} gives for a worker who over-trusts. But the proposition compares an over-trusting user with a calibrated one, not with an unaided control, and a calibrated worker facing low stakes (a large $\qd$, plausible in an experiment) would also check little. The published data cannot tell these readings apart. An experiment that varied the stakes, or told participants the model's error rate on the task, could.

\paragraph{The price of not checking.} By \cref{prop:optimal}, a worker should stop checking only when $q\le\qd$. But $q$ is not printed on the model; it has to be estimated from audited drafts. After $k$ audited drafts with no failures, the most one can claim at 95\% confidence is $q\le 1-0.05^{1/k}$ \citep{clopper1934use}, so certifying $q\le\qd$ at 95\% confidence takes $k\ge\ln 20/{-\ln(1-\qd)}\approx 3/\qd$ clean audits \citep{hanley1983nothing}. In the example above that is 352 drafts, all checked and all clean, on this kind of task. Autonomy is earned by checking. It is not earned on a leaderboard.

\paragraph{Neighbours.} Besides Koopman, the closest model is \citet{huang2026delegation}, in which a worker chooses whether to delegate and how hard to verify, with detection curves such as $1-e^{-\alpha s}$, and optimistic beliefs cause over-delegation that degrades quality. They model utility rather than minutes and do not derive the closed forms of \cref{prop:overtrust}. Treating a human monitor as a signal detector goes back at least to \citet{sorkin1985systems}, and \citet{vasconcelos2023explanations} show experimentally that people trade the cost of checking against the stakes. \Cref{sec:borrowed} itemises what is new.

\paragraph{Predictions.} If the model is right: (P1) measured checking time falls linearly in the log of the model's error rate, not linearly in its accuracy; (P2) among calibrated workers above $\qd$, escaped-defect rates do not improve when the model improves; (P3) making outputs easier to verify (sources, tests, diffs) lowers escapes and total minutes, while accuracy gains alone mainly shorten the clock; (P4) inducing over-trust produces work that is faster and worse, never faster and better. P3 and P4 can be tested in an ordinary two-arm study with an audit; P1 and P2 need several models, or an error rate the experimenter controls. \Cref{sec:reviewers} tests P1 and P4 on AI reviewers, with the error rate under control.

\paragraph{Robustness.} The closed forms depend on A2 and A5. Under Weibull detection with shape $0.5$ or $2$, the log law fails: optimal checking time is no longer linear in $\ln q$, and the calibrated escape rate drifts. Over-trust is still weakly faster and worse in all 12 comparisons I ran, strictly in 11, as part (i) guarantees. Without A5, if the worker stops at the first defect found, checking time follows the log-odds of $q$ rather than its log, and escapes become $\qd(1-q)/(1-\qd)$: $0.84\%$, $0.77\%$ and $0.60\%$ at $q$ of 2\%, 10\% and 30\%, against $\qd=0.85\%$. Invariance then holds only approximately, and better models let slightly more through, not less (\cref{sec:sims}).

% ─────────────────────────────────────────────────────────────────────────────
\section{What a study can identify}
\label{sec:ident}

\paragraph{Three summaries.} Studies report ``the speedup'' as one of three numbers, often without saying which. For tasks timed both ways, write $\rho=T_A/T_B$. The \emph{mean of ratios} is $\E[\rho]$; the \emph{geometric mean ratio} is $\exp\E[\ln\rho]$, which is what a regression on log time estimates; the \emph{ratio of means} is $\E[T_A]/\E[T_B]$, which is $1-\tau$.

\begin{proposition}[The summaries disagree]\label{prop:summaries}
(i) The mean of ratios is at least the geometric mean ratio, with equality only if $\rho$ is constant. (ii) The ratio of means equals $\E[w\rho]$ with weights $w=T_B/\E[T_B]$, so it falls below the mean of ratios when the tool saves proportionally more on long tasks. (iii) They can disagree in sign. If $\ln T_B\sim N(\mu,1)$ and $\ln T_A=\ln T_B+0.1-0.3(\ln T_B-\mu)+v$ with $v\sim N(0,0.3^2)$ independent of $T_B$, the geometric mean ratio says $+10.5\%$, the mean of ratios $+20.9\%$, and the ratio of means $-10.4\%$.
\end{proposition}

In the example the tool costs time on the typical task and saves time on the hours that matter. I recommend the ratio of means, because it answers the question managers ask. With covariates, Poisson pseudo-maximum likelihood (PPML) estimates it consistently, while log-linear regression is biased for it under heteroskedasticity \citep{santossilva2006log}. None of this is new. Computer architects have argued about which mean to report for benchmark speedups since at least \citet{fleming1986lie}, and METR's own appendix reports both summaries \citep{becker2025metr}. The point is only that a headline has to name its summary.

\begin{proposition}[What a single-condition design identifies]\label{prop:ident}
Suppose each task is done once, in one arm, with arms randomized. Then (i) the distributions of $T_A$ and $T_B$ are identified, and with them $\tau$, $\delta$ and any contrast of means; (ii) the share of tasks the tool helps on, $\Pr(T_A<T_B)$, is identified only up to the Makarov bounds \citep{makarov1981estimates}; and (iii) task-level time saved is not identified at all, and the mean of ratios only up to bounds.
\end{proposition}

Part (iii) is the fundamental problem of causal inference \citep{holland1986statistics}; part (ii) is its standard partial-identification answer \citep{heckman1997making,manski2003partial}. They matter here because per-task speedups and ``the share of tasks AI helps on'' are exactly what frontier statistics, TSB's included, try to report. On METR's data, the Makarov bounds on that share are 3\% to 82\% (\cref{sec:record}). Part (iii) also bears on the choice between summaries: in most studies the mean of ratios can only be bounded, not estimated.

\paragraph{Frontiers and the winner's curse.} ``The share of tasks a model handles reliably'' is popular and treacherous. Suppose each task gets $k$ audited drafts, and a task is labelled reliable if at least a fraction $r^\star$ pass. Two things go wrong. First, averaging over the selected tasks on the same drafts that selected them overstates their reliability. If every task has true reliability $0.8$, with $k=10$ and $r^\star=0.9$, then 37.6\% of tasks are labelled reliable and their reliability is reported as $0.93$ (\cref{tab:sims}). Selecting on half the drafts and estimating on the other half removes the bias \citep{athey2016recursive}; \citet{andrews2024inference} give formal inference on winners. Second, identification. With $k$ drafts per task, the data pin down only the first $k$ moments of the distribution of reliability across tasks, since the success-count distribution is a linear image of them. So the share above $r^\star$ is generally only partially identified, even with infinitely many tasks. Under Beta(8,\,2) heterogeneity with $k=10$, it lies anywhere in $[8\%,41\%]$ (truth $22.5\%$). More drafts per task narrow the set; I report only $k=10$, because beyond it the linear programme that computes the bounds is numerically unreliable. A parametric empirical-Bayes fit recovers it when the parametric family is right ($22.9\%$) and misses when it is wrong ($40\%$ against a true $50\%$ for a two-point mixture). The honest options are to fix the frontier before seeing the data, as \citet{dellacqua2026jagged} did by choosing the task in advance; to report the identified set; or to state the model.

\paragraph{Clustering and power.} Workers contribute many tasks, so tasks are not independent. When the tool's effect varies by worker, a bootstrap over tasks covers a nominal 95\% interval 88\% of the time; a cluster bootstrap with 16 workers, 91\%; a leave-one-worker-out jackknife with $t_{G-1}$ critical values, 95\% (\cref{tab:sims}; see \citealp{cameron2008bootstrap,abadie2023should}). For power: METR's design gives a standard error of $0.080$ on log time from 246 issues by 16 developers, so its minimum detectable effect at 80\% power is about 25\%. Detecting a 10\% effect (a time ratio of 1.1) at the same noise would take about 1{,}350 issues, with developers added in proportion.

% ─────────────────────────────────────────────────────────────────────────────
\section{Checks by simulation}
\label{sec:sims}

The main estimators and closed forms above were checked by simulation or against an exact solver; \cref{tab:sims} summarises. These checks verify algebra and estimators. They do not verify the world.

\begin{table}[t]
\centering\small
\caption{Simulation checks (seeds fixed; \texttt{reproduce.py}, about two minutes).}
\label{tab:sims}
\begin{tabularx}{\linewidth}{@{}>{\raggedright\arraybackslash}p{3.1cm}X@{}}
\toprule
Check & Result \\
\midrule
TSB's adjustment on observed times & With completion times that include repairs, TSB's reliability-adjusted score minus true time saved is $-0.100$, $-0.298$, $-0.699$ for draft error rates $0.1$, $0.3$, $0.7$: minus the realised draft error rate, exactly. \\
Frontier, winner's curse & 200 tasks, all with true reliability $0.8$; $k=10$, $r^\star=0.9$. Labelled reliable: 37.5\% (exact: 37.6\%). Their reliability, same drafts: $0.929$. Cross-fitted: $0.800$. \\
Frontier, identification & Beta(8,\,2) reliabilities, infinitely many tasks, $k=10$. Identified set for the share $\ge 0.9$: $[0.082,0.407]$, stable to $0.001$ across grids of 2{,}001 to 8{,}001 points; truth $0.225$. \\
Makarov bounds & Closed form matches an exact transport linear programme over all couplings, on 20 random data sets with ties (unit tests) and on METR's data (\texttt{metr\_reanalysis.json}). \\
Interval coverage & 16 workers $\times$ 15 tasks, worker-level effect s.d.\ $0.3$, 1{,}000 replications. Nominal 95\%: task bootstrap 87.8\%, cluster bootstrap 91.0\%, jackknife-$t$ 95.0\%. With 60 workers: 89.0\%, 93.6\%, 94.9\%. \\
Three summaries & \Cref{prop:summaries}(iii) with $10^6$ draws: geometric mean ratio $1.105$, mean of ratios $1.209$, ratio of means $0.896$, each within $0.001$ of its closed form. \\
Checking model & Event-by-event simulation, $2\times10^6$ drafts per cell, common random numbers: clock time, escapes and the total-minutes loss $\Delta(m)$ match the closed forms, the loss within $0.08$ minutes in every cell with $q>\max(1,m)\qd$. \\
Non-exponential detection & Weibull shape $0.5$ and $2$ (scale $\kappa$): over-trust weakly faster and worse in 12 of 12 comparisons, strictly in 11; the log law fails. \\
Stopping at the first defect & Without A5, escapes equal $\qd(1-q)/(1-\qd)$ at the numerically optimal check time: $0.84\%$, $0.77\%$, $0.60\%$, $0.34\%$ at $q$ of 2, 10, 30, 60\%. \\
\bottomrule
\end{tabularx}
\end{table}

% ─────────────────────────────────────────────────────────────────────────────
\section{The record, re-read}
\label{sec:record}

\subsection{METR, replicated}

METR released its per-issue data \citep{metr2025data}. Re-running its specification (log time on treatment and the developer's pre-randomization forecast) reproduces its numbers exactly: $+18.8\%$, with a developer-clustered 95\% interval of $[+1.6\%,+39.0\%]$. \Cref{tab:metr} reads the same 246 issues two ways (two estimands), each with and without the forecast adjustment.

\begin{table}[t]
\centering\small
\caption{One data set, several headlines: METR's early-2025 study \citep{becker2025metr}, 246 issues, 16 developers. Positive means more time with AI. Intervals are leave-one-developer-out jackknife with $t_{15}$ critical values unless marked.}
\label{tab:metr}
\begin{tabular}{@{}lll@{}}
\toprule
Estimand & Forecast-adjusted & Unadjusted \\
\midrule
Geometric mean ratio & $+18.8\%$ $[-0.2,\ +41.6]$ & $+23.0\%$ $[+0.8,\ +50.0]$ \\
 & \quad METR: $[+1.6,\ +39.0]^{a}$ & \\
Ratio of means & $+24.0\%$ $[+0.4,\ +53.2]$ (PPML) & $+33.7\%$ $[+8.7,\ +64.4]$ \\
\midrule
\multicolumn{3}{@{}l}{Share of issues on which AI helped: Makarov bounds $[2.6\%,\ 82.1\%]^{b}$} \\
\multicolumn{3}{@{}l}{Post-review fix time, AI minus no AI: $+5.2$ min, cluster bootstrap $[-0.6,\ +12.1]^{c}$} \\
\bottomrule
\multicolumn{3}{@{}p{0.95\linewidth}@{}}{\footnotesize $^{a}$METR's published developer-clustered interval (normal critical values), reproduced exactly. $^{b}$Plug-in bounds from the two arms' distributions of log(time/forecast); sampling error not shown, and conditioning on the forecast could only narrow them. $^{c}$On the 230 issues where post-review time was recorded.}
\end{tabular}
\end{table}

Under randomization, the adjusted and unadjusted columns estimate the same thing. They differ here because AI-allowed issues happened to be forecast longer, by 9\% on average (116 against 106 minutes) and by 4\% in geometric mean (91 against 87), and the adjusted column corrects for that. The real choice is between rows. ``AI made developers 19\% slower'' and ``AI cost developers 24\% more hours'' are both supported by the same data, and a headline that does not name its summary is choosing between them silently. METR chose a defensible estimator and reported the unadjusted 34\% itself; the point is not that they chose wrong. The data give little evidence that the gap between rows is about task length: the interaction between treatment and log forecast is $0.06$, with interval $[-0.11,0.23]$. It is about spread. AI-allowed times are more dispersed (residual s.d.\ $0.66$ against $0.60$ on the log scale), and under log-normality that alone turns $18.8\%$ into about $23\%$.

Two more things show up. With 16 developers, inference leans on the small-sample method: the jackknife-$t$ interval just includes zero, while METR's interval does not. (The same method covered at 95\% in my simulation with 16 workers, for the ratio of means.) METR's appendix shows other corrections, including CR2 and CR3 cluster-robust errors and a two-stage bootstrap \citep[Fig.~15]{becker2025metr}; I note the sensitivity, not a verdict. And the Makarov row says something no average can: these data are consistent with AI having helped on most issues, or on almost none.

The closest thing to an escaped-defect measure in the data is the time spent fixing a pull request after review, which is work a reviewer sent back. On the 230 issues where it was recorded, it was 14.9 minutes per AI-allowed issue against 9.7 without. The difference is not significant. Its sign is the one \cref{prop:overtrust} would predict, which is not evidence for it.

METR's late-2025 follow-up \citep{metr2026uplift} also shows a gap between summaries, here in the other direction: for returning developers, $-18\%$ by log-time OLS and $-25\%$ by PPML. With nine returning developers, and the selection METR warns about, I draw nothing else from it.

\subsection{GDPval, bounded}

GDPval's paper models an expert who samples a draft, reviews it, and redoes the task if the draft is unsatisfactory \citep[App.~A.2.1]{patwardhan2025gdpval}. On its gold subset an expert takes 404 minutes and review takes 109; GPT-5's drafts take about 4.5 minutes ($404/90$) and are rated better than the expert's own work 39\% of the time. Assuming a failed draft is redone from scratch, the paper gets $1.12\times$ (one try) and $1.39\times$ (repeated tries).

The fix cost is the assumption doing the work. Let a failed draft cost anywhere from nothing (it is trivially fixed) to a full redo, and let the acceptance rate be at least the win rate (a draft that beats the expert's own is acceptable). The one-try speedup is then identified only up to
\[
  \left[\frac{404}{4.5+109+0.61\times 404},\ \frac{404}{4.5+109}\right]=[1.12\times,\ 3.56\times].
\]
The upper end assumes nothing about the quality of the model at all. It is \cref{prop:ceiling} applied to GDPval's own workflow, in which every output is reviewed, with GDPval's grading time standing in for review time: review takes 27\% of the expert's time, so with GPT-5's generation time this workflow cannot be more than $3.56\times$ faster, and with instant generation not more than $404/109\approx 3.7\times$, unless review itself gets faster. The 100$\times$ in the announcement is model time. For this workflow the deployed number lies between $1.1\times$ and $3.6\times$, and the paper's own estimate sits at the bottom of that range. GDPval also assumes review is perfect. Under \eqref{eq:search}, any finite review misses something, so the escaped-defect side of the ledger is unmeasured.

\subsection{The jagged frontier, priced}

On the task chosen to lie outside the frontier, the control group was correct 84.5\% of the time and took 37.7 minutes on average. Consultants given GPT-4 with an overview of prompting were correct 60.0\% of the time and took 26.2 minutes; those given GPT-4 alone, 70.6\% and 30.9 minutes \citep{dellacqua2026jagged}. That is $\tau=30\%$ and $18\%$, and $\delta=+24.5$ and $+13.9$ percentage points. The break-even defect costs are
\[
  \lambda^\star=\frac{11.5\ \text{min}}{0.245}\approx 47\ \text{min},\qquad \frac{6.8\ \text{min}}{0.139}\approx 49\ \text{min}.
\]
Assistance paid on this task only if a wrong answer could be caught and undone later for less than about 48 minutes of a consultant's time, about 1.2 to 1.3 times the control group's time on the whole task. I doubt many clients would sign that.

\subsection{The rest of the record}

\Cref{tab:record} restates the main published results in the form of \cref{sec:estimand}, using only what each source reports. The quality measures differ across studies, so the table compares regions, not magnitudes. Most results are dominant or dominated. The trade-offs are where the interesting decisions are, and they run both ways: the consultants outside the frontier bought time with errors, and the physicians in \citet{goh2025gpt4} bought better reasoning with time. Only for the first can $\lambda^\star$ be computed from what was published, because only there is quality reported as an error rate.

\begin{table}[t]
\centering\small
\caption{Published results as (time, quality) pairs. Regions, as defined in \cref{sec:estimand}, use point estimates; ``n.s.'' marks a quality difference that was not statistically significant. Sources report different quality measures, so compare regions, not magnitudes.}
\label{tab:record}
\begin{tabularx}{\linewidth}{@{}>{\raggedright\arraybackslash}p{3.9cm}>{\raggedright\arraybackslash}X>{\raggedright\arraybackslash}X>{\raggedright\arraybackslash}p{2.3cm}@{}}
\toprule
Study & Time & Quality & Region \\
\midrule
\citet{peng2023copilot}, Copilot & $-55.8\%$ time $[-89,-21]$ & task success $+7$ pp, n.s. & dominant \\
\citet{noy2023experimental}, writing & $-40\%$ time & grade $+18\%$ & dominant \\
\citet{dellacqua2026jagged}, inside & $-25.1\%$ time & quality score $+30$--$34\%$ & dominant \\
\citet{dellacqua2026jagged}, outside & $-30\%$ / $-18\%$ time & wrong answers $+24.5$ / $+13.9$ pp & trade-off; $\lambda^\star\approx 48$ min \\
\citet{brynjolfsson2025generative}, less-skilled & $+30\%$ resolutions/hour & speed and quality both up & dominant \\
\citet{brynjolfsson2025generative}, most-skilled & small or no gain & small decline & trade-off or dominated \\
\citet{goh2025gpt4}, physicians & $+119$ s per case $[17,221]$ & reasoning score $+6.5\%$ $[2.7,10.2]$ & trade-off (slower, better) \\
\citet{becker2025metr} & $+19\%$ time $[2,39]$ & post-review fixes $+5$ min, n.s. & dominated \\
\citet{patwardhan2025gdpval}, GPT-5 & $1.12\times$ to $3.56\times$ & not measured after review & unknown \\
\bottomrule
\end{tabularx}
\end{table}

% ─────────────────────────────────────────────────────────────────────────────
\section{A test on AI reviewers}
\label{sec:reviewers}

Predictions P1--P4 concern anyone who checks someone else's draft, including an AI reviewer checking an AI draft. That is a checker already deployed at scale, and one that is cheap to study under control: the experimenter can set the error rate and tell the reviewer what it is. This section tests P1 and P4 on one reviewer model. It says nothing directly about people.

\paragraph{Design.} The task is the brief of \cref{sec:intro}: revenue, net income and operating margin for three companies for the quarter ended June 30, 2026, from 10-Qs filed from July 1, 2026, after the models' training data were collected \citep{sec2026edgar}. Ground truth is exact: for 30 of 30 sampled companies, all twelve extracted figures appear on the printed statement. A deterministic auditor scores every final brief. A draft has no defect or one seeded wrong figure: a stale period, a year-to-date figure, another company's figure, transposed digits, or the wrong numerator. (FinVerBench also seeds errors in XBRL statements, as percentage changes to one line; \citealp{panda2026finverbench}.) The reviewer, \texttt{claude-sonnet-5-5}, has one tool, \texttt{lookup}, which shows one line of one company's statement. Its lookups, counted by the tool server, are its checking effort, and nine cover every line. Its prompt prices a lookup at 3 minutes and an escaped error at 10 hours, so checking every line pays exactly when the stated error rate exceeds $q_{bb}=27/600=4.5\%$. Pipeline drafts come from \texttt{claude-haiku-5-5}; the unaided baseline is \texttt{claude-sonnet-5-5} at high effort. Each agent is a separate, isolated Claude Code session, 540 in the analysed run. Hypotheses, design and analysis code were fixed after a 66-trial pilot on other filings and frozen in a commit to the repository before the main run; the freeze was not lodged with an outside registry. One instrument fault surfaced mid-run: three company names held a no-break space, so lookups of them failed. It was fixed, and the 23 affected trials were re-run. Intervals are cluster bootstraps over 40 filings packets; the three primary tests are Holm-corrected.

\begin{table}[t]
\centering\small
\caption{Pre-specified hypotheses and results, AI reviewers (\texttt{claude-sonnet-5-5}) checking AI drafts. Brackets are 95\% cluster-bootstrap intervals. $p$-values are bootstrap, Holm-corrected across the three primary tests (2{,}000 resamples, so $p<.001$ is the floor).}
\label{tab:reviewers}
\begin{tabularx}{\linewidth}{@{}>{\raggedright\arraybackslash}p{3.9cm}>{\raggedright\arraybackslash}X>{\raggedright\arraybackslash}p{2.35cm}@{}}
\toprule
Hypothesis (experiment, $n$) & Estimate & Verdict \\
\midrule
H1. Detection is concave in effort, A2 (E1; 25 defective drafts per budget) & Caught 0, 32, 48, 100\% with 0, 1, 3, 9 lookups. Exponential fit preferred to linear (AIC 75.0 vs 77.8; $\kappa=3.1$ lookups) & confirmed \\
\textbf{H2.} Effort rises with $\ln q$, P1 (E2; 120 reviews, $q$ stated truthfully) & $+2.2$ lookups $[1.8,\ 2.5]$ per unit of $\ln q$; $p<.001$ & \textbf{confirmed} \\
H2b. Effort is graded, not all-or-nothing (E2) & 96\% $[91,\ 99]$ of reviews made 0 or 9 lookups; the switch lies between $q=2\%$ and $5\%$ & all-or-nothing \\
H3. Escapes flat in $q$ where checked, P2 (E2, $q\ge 10\%$) & No escapes at any of these levels; slope $0$ $[0,\ 0]$ & estimate only \\
\textbf{H4a.} Over-trust checks less, P4 (E3; 40 paired drafts, true $q=30\%$, told $3\%$) & $-7.2$ lookups $[-8.1,\ -6.0]$; $p<.001$ & \textbf{confirmed} \\
\textbf{H4b.} \ldots and lets more through, P4 (E3) & Escapes $22.5\%$ vs $0\%$: $+22.5$ points $[10,\ 35]$; $p<.001$ & \textbf{confirmed} \\
H4c. \ldots beyond what its lower effort predicts (E3 with E1's curve) & Excess misses $-0.03$ $[-0.15,\ 0.10]$; every review that looked up a line exposing the wrong figure fixed it (20 of 20) & no: all through effort \\
Told 3\%, plus ``treat track-record claims as unverified'' (E3) & $-7.0$ lookups $[-8.0,\ -5.9]$; escapes $+17.5$ points $[5,\ 30]$ & instruction did not help \\
H6. Haiku drafts plus review vs Sonnet alone (E5; 20 packets) & No escapes in either arm ($\delta=0$). Pipeline $2.2\times$ the cost $[2.15,\ 2.23]$ and $3.2\times$ the time $[2.9,\ 3.5]$ & dominated \\
H7. Winner's curse in ``share reliable'' (E6; 120 drafts) & All 120 drafts fully correct; nothing to select on & not testable \\
\bottomrule
\end{tabularx}
\end{table}

\begin{figure}[t]
  \centering
  \includegraphics[width=\linewidth]{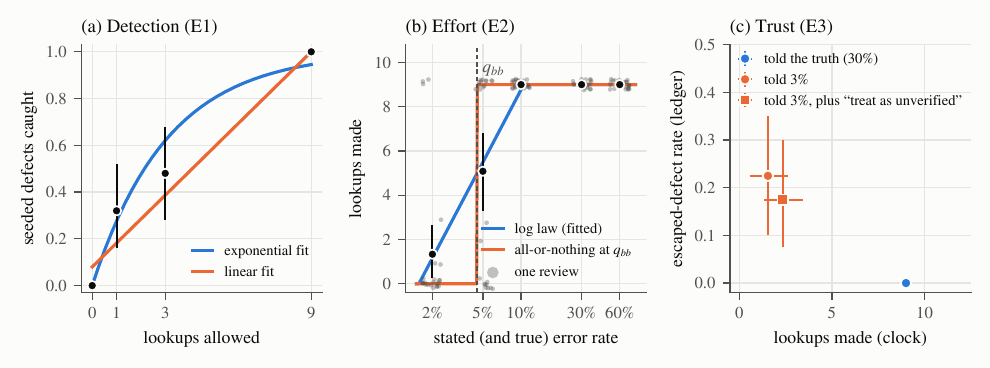}
  \caption{AI reviewers checking AI drafts. (a) Seeded defects caught under a fixed number of lookups, with both pre-specified fits; nine lookups check every line. (b) Lookups chosen when told the drafter's true error rate: each grey dot is one review (jittered), black dots are means. The mean rises with $\ln q$, but almost every review checks nothing or everything, switching near the break-even rate $q_{bb}$ that the prompt's costs imply. (c) Same drafts, true error rate 30\%: a reviewer told 3\% looks less and lets more through. The two ``told 3\%'' arms are each shifted $0.3$ lookups sideways so both are visible (their means are $1.85$ and $2.05$). Bars are 95\% cluster-bootstrap intervals. Generated from the run's logs by \texttt{reviewlab/figure.py}.}
  \label{fig:reviewers}
\end{figure}

\paragraph{The reviewer does its prompt's arithmetic.} Told the truth, the reviewer checked more as the error rate rose (H2), but not as \cref{prop:optimal} says (\cref{tab:reviewers,fig:reviewers}). Almost every review checked nothing or everything, switching near where the prompt's costs break even. At 2\%, three in four checked nothing. At 5\%, where a full check (27 minutes) and none (an expected 30) nearly tie, 54\% checked every line. From 10\% up, all did. Every review at 2\% that made one lookup or none wrote the comparison out (20 of 20). The mean follows $\ln q$; no review does. All-or-nothing is optimal only if partial checks are worth nothing, and here they were not. With one lookup the reviewer caught a third of seeded defects, landing on the defective line twice as often as a random lookup would (32\% against 15\%), and detection is concave (H1). So the reviewer under-uses partial checking, though in this task that costs little. On an exploratory reading of the E1 rates, one lookup beats both corners by under a minute per draft at 2\% and by about four at 5\%. Search here is also finite: nine lookups caught every seeded defect, so wherever the reviewer checked, nothing escaped. The escape floor $\qd$ of \cref{prop:optimal}(ii) assumes a search that never runs out.

\paragraph{Over-trust is faster and worse.} \Cref{prop:overtrust}(i), which holds for any detection curve, predicts that a reviewer who over-rates the drafter checks less and lets more through. Told 3\% when the truth was 30\%, 29 of 40 reviews made no lookup, against none of 40 when told the truth. A wrong figure reached the client in 22.5\% of briefs instead of none. In the prompt's units, over-trust saved 21 minutes of checking per draft and cost $0.225\times 600=135$ minutes downstream: a break-even defect cost of about 95 minutes, against the 600 the prompt names. The effect ran entirely through effort. Across the three arms, every review that looked up a line exposing the wrong figure corrected it (20 of 20), and all 16 escapes came from reviews that never looked at such a line; the pre-specified test agrees (H4c). The reviewer did not accept more at a given effort; it looked less. Telling it to treat the stated record as unverified did not help: 26 of 40 reviews in that arm called the 3\% figure unverified and then checked nothing. That matches LLM judges, whose anchoring on prior numerical scores survived a ``disregard'' warning \citep{kapetanovic2026anchoring}. In LLM code review, redacting misleading metadata, rather than warning against it, restored detection \citep{alexopoulos2026contextual}. Faster and more often wrong is also what \citet{dellacqua2026jagged} found outside the frontier. That is a parallel, not a replication.

\paragraph{What the review was worth.} The Haiku drafter made no natural errors in 140 briefs. So H7 had nothing to select on, and no arm of the pipeline comparison let anything through. The reviewer, told nothing about the drafter, checked every line of all 20 drafts and corrected nothing. The pipeline cost 2.2 times as much as Sonnet alone and took 3.2 times as long: dominated. The certification rule of \cref{sec:model} prices that review. Counted as independent audits (generously, since they cluster on 20 packets), 140 clean briefs bound the error rate below 2.1\% at 95\% confidence. That is under the prompt's break-even 4.5\%, which takes about $3/q_{bb}\approx 67$ clean audits to certify. At these stakes the review could have been skipped, but only because someone counted.

\paragraph{Limits.} The study tests one reviewer model, with its costs written into its prompt. A reviewer that does the arithmetic it is given is following instructions; that is how such pipelines are configured, and it is what was tested. Seeded defects can overstate detection \citep{kumar2026bigger}, and the planned check against natural errors found none to check. Another company's figure, which leaves a duplicate in the draft, was caught most often: 89\% across budgets of one to nine lookups, against 44--67\% for the other kinds. A defect counts as caught only if it is corrected to the true value. With exactly $\mathrm{round}(qn)$ defective drafts, E2 had none at 2\% and one at 5\%, so it is silent on escapes at low error rates and on the low-prevalence effect of human search \citep{wolfe2005rare}. And the study is small: 540 sessions, plus 24 superseded, on one author's subscription, worth \$9.27 at list price and powered only for large effects. A full-scale design is pre-specified in the repository.

% ─────────────────────────────────────────────────────────────────────────────
\section{A reporting standard}
\label{sec:standard}

\Cref{tab:checklist} is the checklist I would want every time-saved claim to meet. Its form follows \citet{wei2025human}. None of it is expensive; most of it is bookkeeping.

\begin{table}[t]
\centering\small
\caption{Reporting checklist for claims that AI saves time.}
\label{tab:checklist}
\begin{tabularx}{\linewidth}{@{}rX@{}}
\toprule
1 & Name the summary (ratio of means, geometric mean ratio, or other), and the populations of tasks and workers it averages over. \\
2 & Use observed times from both arms. Self-reports are not a substitute: METR's developers misjudged their own speedup by about 40 points. \\
3 & Audit final deliverables blind, in both arms, and report $\delta$ with an interval. \\
4 & When $\tau$ and $\delta$ have the same sign, report $\lambda^\star$. \\
5 & If instrumented, report checking time and repair events \citep[coded as in][]{mozannar2024reading}. \\
6 & Report inference cost in money, in its own column. \\
7 & Cluster inference by worker; with few workers, use a small-sample method such as the jackknife-$t$. \\
8 & Define frontiers and subgroups before the data, or cross-fit; report identified sets for ``share helped'' and ``share reliable''. \\
9 & Back any ``no review needed'' claim with roughly $3/\qd$ clean audits on that kind of task, where $\qd$ is the error rate below which skipping review is acceptable. \\
10 & Pre-register the analysis and state the power. \\
\bottomrule
\end{tabularx}
\end{table}

% ─────────────────────────────────────────────────────────────────────────────
\section{What is borrowed, and what is new}
\label{sec:borrowed}

I searched arXiv, SSRN, NBER, PMLR and ACM venues through October 2026. \Cref{tab:borrowed} lists what, as far as that search found, this paper adds, next to the closest prior work and the difference. The reader can judge the distance.

\begin{table}[t]
\centering\small
\caption{Contributions, and the nearest thing that already existed.}
\label{tab:borrowed}
\begin{tabularx}{\linewidth}{@{}>{\raggedright\arraybackslash}p{5.2cm}>{\raggedright\arraybackslash}X@{}}
\toprule
Contribution & Closest prior work, and what is new \\
\midrule
Break-even defect cost $\lambda^\star$ and the three regions & The incremental cost-effectiveness ratio and the cost-effectiveness plane \citep{black1990ce}; team utility with a cost per mistake \citep{bansal2021most}; verification friction \citep{kamara2026verification}. New is only the application to time-saved claims. \\
$\varepsilon^\star=\min(q,\qd)$; checking time linear in $\ln q$ & Koopman's optimal search effort, of which this is the one-cell case \citep{koopman1957search}; effort-optimal verification \citep{huang2026delegation}. New is the reading for AI: better models buy speed, not quality. \\
Over-trust loss $\kappa(m-1-\ln m)$, its asymmetry, the clock/ledger split & \citet{huang2026delegation}: optimistic beliefs cause over-delegation, shown qualitatively and in utility rather than minutes. \\
Certification cost $\approx 3/\qd$ & The rule of three \citep{hanley1983nothing}; new is tying it to the checking threshold. \\
Identified set for ``share of tasks reliable'' & Standard moment arguments for binomial mixtures; new is the application to frontier statistics. \\
Makarov bounds, PPML and jackknife-$t$ on METR's data & METR's appendix, which reports log-time OLS, the raw ratio of means and other small-sample corrections \citep{becker2025metr}. \\
Bounds on GDPval's deployed speedup & GDPval's point estimates, which fix the repair cost at a full redo. \\
$\lambda^\star$ for the outside-frontier task & \citet{dellacqua2026jagged}, who report time and correctness separately. \\
A checklist for time-saved claims & \citet{wei2025human}, for human baselines in general. \\
AI reviewers' effort against a stated error rate; over-trust split into effort and acceptance & Reassuring framing cuts LLM code reviewers' detection \citep{alexopoulos2026contextual}; prior scores anchor LLM judges \citep{kapetanovic2026anchoring}; errors seeded in XBRL statements \citep{panda2026finverbench}. New: a numeric error rate set by the experimenter, effort counted as logged lookups, the all-or-nothing policy, the split of over-trust's escapes into effort and acceptance, and an independent auditor on filings made after the models' training data were collected. \\
\bottomrule
\end{tabularx}
\end{table}

Everything else is used as found: the cost-effectiveness plane \citep{black1990ce}; optimal search \citep{koopman1946search,koopman1957search,stone1975theory}; Amdahl's bound and its application to AI-assisted work \citep{amdahl1967validity,patwardhan2025gdpval,liang2026novelty}; monotone comparative statics \citep{topkis1978minimizing,milgrom1994monotone}; monitors as signal detectors \citep{sorkin1985systems}; checking as a cost--benefit choice, and verifiability \citep{vasconcelos2023explanations,fok2024verifiability}; the cost of a correct output \citep{erol2026costofpass}; activity coding \citep{mozannar2024reading}; PPML \citep{santossilva2006log}; partial identification \citep{makarov1981estimates,heckman1997making,manski2003partial}; honest estimation and inference on winners \citep{athey2016recursive,andrews2024inference}; clustered inference \citep{cameron2008bootstrap,abadie2023should}; zero-failure bounds \citep{clopper1934use,hanley1983nothing}; and, underneath all of it, the pairing of speed with quality and the jagged frontier itself \citep{dellacqua2026jagged}.

% ─────────────────────────────────────────────────────────────────────────────
\section{Limitations}
\label{sec:limits}

The model is stylised. It allows one defect per draft, exponential search, a fixed checking budget, risk-neutral workers and no false alarms; \cref{sec:model} says what survives when the search law or the budget is relaxed. A false alarm adds repair time to the clock and nothing to escapes; it shifts the closed forms but not \cref{prop:overtrust}(i). Real checking finds defects in batches, and real repairs can leave something behind.

The audit defines $\delta$. Audits miss things too. An audit that misses the same share of defects in both arms shrinks $\delta$ toward zero; it cannot manufacture a difference. An audit that is better at catching one kind of defect than another can.

Nothing here measures $c$, the downstream cost of a defect. $\lambda^\star$ is built so that a benchmark need not, but someone deciding whether to deploy has to compare it with their own $c$.

The re-analysis uses published numbers, and their quality measures are not comparable across studies; I compare regions only. The Makarov bounds on METR's data are plug-in estimates without sampling error. Predictions P1 and P4 are tested only on AI reviewers: one model, told its costs, on one kind of task, with seeded defects, in a study powered for large effects (\cref{sec:reviewers}). P2 and P3 are untested. The reviewer and the drafters are Claude models, and Claude helped prepare this paper. I collected no data on people.

% ─────────────────────────────────────────────────────────────────────────────
\section{Conclusion}
\label{sec:conclusion}

A time-saved number is a claim about two worlds, and it is only as good as the bookkeeping in the second. The stopwatch says how fast the work went. It cannot say what the work let through, and wherever a model is good enough to use but not good enough to use unchecked, that is where the time goes: into checking, or into the defects that checking would have caught. In the one experiment here, an AI reviewer told its drafter was more reliable than it was did exactly what the ledger view warns about: it saved minutes on the clock and let wrong figures through to the client. Name the summary. Audit both arms. Report the break-even defect cost, and let the reader price the defect.

Gilbreth, to his credit, counted the bandages.

\vspace{10pt}
\noindent\textbf{Code and data.} Every simulated or re-analysed number in \cref{sec:model,sec:ident,sec:sims,sec:record} is regenerated by the script \texttt{reproduce.py}, and every published number it restates is listed with its source in \texttt{reanalysis.py}. Both, with 46 unit tests, are in the \texttt{v2} folder of \url{https://github.com/solomonshalom/tsb-bench}. METR's data are downloaded from its repositories at pinned commits and checked against SHA-256 digests, not redistributed. Corrections to TSB are listed, with tests, in \texttt{NOTES.md}. The experiment of \cref{sec:reviewers} is in \texttt{v2/experiments}: the SEC data snapshot (US-government public data) with its SHA-256 digest, the pre-registration and its freeze record, the record of deviations, the harness with 35 unit tests, every trial's log and transcript, and the analysis that produces \cref{tab:reviewers} and \cref{fig:reviewers}.

\vspace{4pt}
\noindent\textbf{Author note.} I am an independent researcher, and this work received no external funding. I work on Rōmy, which builds AI software; nothing in this paper evaluates it.

\vspace{4pt}
\noindent\textbf{Use of AI assistance.} This paper was prepared with substantial help from Anthropic's Claude (Opus 5.5): searching the literature, checking each cited number against its primary source, writing the analysis code, the experiment's harness and tests, running the experiment, and drafting prose. The experiment's agents are Claude models too (Haiku 5.5 and Sonnet 5.5), run on my own subscription. I reviewed all of it and am responsible for it.

% ─────────────────────────────────────────────────────────────────────────────
\appendix
\section{Proofs}
\label{app:proofs}

\paragraph{\Cref{prop:ceiling}.} $q\,s(V)R\ge 0$ in \eqref{eq:time}.

\paragraph{\Cref{prop:optimal}.} Up to the constant $P+L$, the worker minimises $J(V)=V+\hat q[sR+(1-s)c]=V+\hat qc-\hat q(c-R)s(V)$. If $\hat q=0$, $V^\star=0$. Otherwise, under \eqref{eq:search}, $J'(V)=1-(\hat q(c-R)/\kappa)e^{-V/\kappa}$ and $J''>0$ since $c>R$. If $\hat q(c-R)\le\kappa$, then $J'\ge 0$ everywhere and $V^\star=0$. Otherwise $J'(V^\star)=0$ at $V^\star=\kappa\ln(\hat q(c-R)/\kappa)=\kappa\ln(\hat q/\qd)$. With $\hat q=q$, the escape rate is $qe^{-V^\star/\kappa}=\qd$ when $q>\qd$, and $q$ otherwise, which gives (ii); (iii) follows from $V^\star(q)-V^\star(q/10)=\kappa\ln 10$. For the remarks on $\kappa$: at the optimum, expected minutes including downstream costs are $J^\star=P+L+\kappa\ln(q/\qd)+qR+\kappa$, and $\partial J^\star/\partial\kappa=\ln(q/\qd)>0$ when $q>\qd$. Clock time alone is $P+L+\kappa\ln(q/\qd)+(q-\qd)R$, whose derivative in $\kappa$ is $\ln(q/\qd)-1-R/(c-R)$, positive only when $q>e^{1+R/(c-R)}\qd$.

\paragraph{\Cref{prop:overtrust}.} (i) A detection curve is the distribution function of the time to find a defect, so it is nondecreasing and right-continuous; hence $-J(V;\hat q)=-V-\hat qc+\hat q(c-R)s(V)$ is upper semicontinuous in $V$. Any maximiser lies in $[0,\hat q(c-R)]$, since beyond that $-J(V)<-\hat qc\le -J(0)$. So the set of maximisers is nonempty and compact, with least and greatest elements. The function has increasing differences in $(V,\hat q)$, because its increment in $\hat q$, $-c+(c-R)s(V)$, is nondecreasing in $V$. By Topkis's theorem the set of maximisers is ascending in $\hat q$, so its least and greatest elements are nondecreasing in $\hat q$. Clock time $V+qs(V)R$ is nondecreasing in $V$ and escapes $q(1-s(V))$ are nonincreasing in $V$, which gives the claim. (ii) Let $V_c=\kappa\ln(q/\qd)$ and $V_m=V_c-\kappa\ln m$, both positive when $q>\max(1,m)\qd$. Then $qe^{-V_c/\kappa}=\qd$ and $qe^{-V_m/\kappa}=m\qd$. Clock time is $P+L+V+q(1-e^{-V/\kappa})R$, so the difference is $\kappa\ln m+(m-1)\qd R$, and escapes go from $\qd$ to $m\qd$. True expected minutes are $P+L+V+qR+q(c-R)e^{-V/\kappa}$, which equals $P+L+V_c+qR+\kappa$ at $V_c$ and $P+L+V_c-\kappa\ln m+qR+m\kappa$ at $V_m$. The difference is $\kappa(m-1-\ln m)$, which is nonnegative because $\ln m\le m-1$. If $\qd<q\le m\qd$, the over-trusting worker sets $V=0$ and pays $qc$; subtracting the calibrated $V_c+qR+\kappa$ gives $\kappa(x-1-\ln x)$ with $x=q/\qd\le m$, which is at most $\Delta(m)$ because $x-1-\ln x$ increases for $x>1$.

\paragraph{\Cref{prop:summaries}.} (i) Jensen's inequality. (ii) $\E[T_A]/\E[T_B]=\E[T_B\rho]/\E[T_B]$. (iii) With $x=\ln T_B\sim N(\mu,\sigma_x^2)$ and $\ln\rho=\beta-\gamma(x-\mu)+v$, $v\sim N(0,\sigma_v^2)$ independent of $x$: the geometric mean ratio is $e^{\beta}$, the mean of ratios is $e^{\beta+(\gamma^2\sigma_x^2+\sigma_v^2)/2}$, and the ratio of means is $e^{\beta+((\gamma^2-2\gamma)\sigma_x^2+\sigma_v^2)/2}$. Substitute $\beta=0.1$, $\sigma_x=1$, $\gamma=0.3$, $\sigma_v=0.3$.

\paragraph{\Cref{prop:ident}.} Randomization identifies the distribution of time in each arm. Any joint distribution with those two marginals fits the data equally well. Over that set, Makarov's bounds are the sharp bounds on $\Pr(T_A<T_B)$. Part (iii) follows because the set contains couplings that give different task-level differences, and different means of their ratios; by the rearrangement inequality those means lie between their values under the comonotone and antitone couplings.

\paragraph{Certification and the identified set.} With $k$ clean audits, $\Pr(\text{all clean}\mid q)=(1-q)^k\le 0.05$ exactly when $q\ge 1-0.05^{1/k}$. For the share of reliable tasks: the success-count probabilities
\[\Pr(S=j)=\int\tbinom{k}{j}\,r^j(1-r)^{k-j}\,dF(r),\qquad j=0,\dots,k,\]
are linear in the first $k$ moments of $F$, so every $F$ with those moments fits the data equally well. The bounds in \cref{sec:ident} are the minimum and maximum of $\Pr_F(r\ge r^\star)$ over such $F$, computed by linear programming (dual simplex, exact equality constraints) on a grid of 4{,}001 points; grids of 2{,}001 and 8{,}001 points give the same bounds to within $0.001$.

% ─────────────────────────────────────────────────────────────────────────────

\end{document}